%% file: ifacconf.tex
\documentclass{ifacconf}

\usepackage{times}
\usepackage{graphicx}      

\usepackage{natbib}        
\usepackage{amsmath}
\usepackage[utf8]{inputenc}
\usepackage{listofitems}
\usepackage{amsmath}
	\usepackage{booktabs}

\graphicspath{{figures/}}

\usepackage{tikz}
\usepackage{verbatim}
\usetikzlibrary{arrows,shapes.geometric,fit,scopes,calc,matrix, positioning,decorations.pathmorphing,fadings,external}
 
\begin{document}
\begin{frontmatter}

\title{Analysis of Numerical Integration in RNN-Based Residuals for Fault Diagnosis of Dynamic Systems} 

\thanks[footnoteinfo]{This work is partially funded by the Swedish Excellence Center ELLIIT.}
\thanks[footnoteinfo]{© 2023 the authors. This work has been accepted to IFAC for publication under a Creative Commons Licence CC-BY-NC-ND}

\author{Arman Mohammadi}, 
\author{Theodor Westny}, 
\author{Daniel Jung}, \textbf{and} 
\author{Mattias Krysander} 

\address{Department of Electrical Engineering, 
   Linköping University, \\ SE-581 83, Linköping, Sweden. \\ e-mail: \{arman.mohammadi, theodor.westny, daniel.jung, mattias.krysander\}@liu.se}


\begin{abstract}
Data-driven modeling and machine learning are widely used to model the behavior of dynamic systems. 
One application is the residual evaluation of technical systems where model predictions are compared with measurement data to create residuals for fault diagnosis applications.
While recurrent neural network models have been shown capable of modeling complex non-linear dynamic systems, they are limited to fixed steps discrete-time simulation.  
Modeling using neural ordinary differential equations, however, 
make it possible to evaluate the state variables at specific times, 
compute gradients when training the model and use standard numerical solvers to explicitly model the underlying dynamic of the time-series data. 
Here, the effect of solver selection on the performance of neural ordinary differential equation residuals during training and evaluation is investigated. 
The paper includes a case study of a heavy-duty truck's after-treatment system to highlight the potential of these techniques for improving fault diagnosis performance. 

\end{abstract}


\end{frontmatter}

\input{introduction}
\input{background}
\input{casestudy}
\input{evaluation}
\input{conclusions}

\bibliography{main}         

\end{document}

%% file: introduction.tex
\section{Introduction}
\vspace{-0.05in}
Fault diagnosis is an important task in ensuring the reliable operation of complex systems. One approach to fault diagnosis involves monitoring system behavior and identifying anomalies through the comparison of model predictions and sensor measurements, which are referred to as residuals. However, an accurate model of the system is required for effective fault diagnosis, 
as model inaccuracies can lead to poor performance in detecting faults. 
Machine learning and data-driven models, such as Recurrent Neural-Networks (RNNs), 
have shown promise as alternatives for developing mathematical models of non-linear dynamic behavior from time-series data \citep{anderson1996comparison}. 
The authors in \cite{pulido2019state} proposed the use of Recurrent Neural Networks (RNNs) derived from physical insights for fault detection. 
These RNNs, here referred to as grey-box RNNs, model the nominal system behavior and are trained using only nominal data to detect abnormal behavior. 
The network structures are designed to resemble the structural properties of model-based residuals, which enables the isolation of unknown fault classes \citep{jung2019isolation}.

Many technical systems can be described by deterministic non-linear dynamic models, such as ordinary differential equations (ODEs). Various numerical solvers with different orders have been developed to find approximate discrete-time solutions to ODEs, see e.g. \citep{ascher1998computer}. 
The order of the solver determines the total accumulated error of a simulation, which is in the order of $\mathcal{O}(T^p)$, where $T$ is the step and $p$ is the order of the solver. Lower-order solvers are typically faster and less computationally expensive but suffer from significant prediction errors. 
They may require smaller simulation time steps and provide less accurate solutions for models with complex dynamics. On the other hand, higher-order solvers can provide more accurate solutions at the cost of computational complexity. 
Adaptive step lengths can be used to balance error tolerance and computation time but there is a risk that the simulation will not finish within a specific time frame if the system is affected by sudden disturbances and faults, e.g. if the solver reduces the step length to fulfill the error tolerances. 

Even though RNNs are black-box models, previous research has shown that certain RNN structures, such as ResNets, can be seen as Euler discretization of continuous dynamic models, see e.g. \cite{lu2018beyond,haber2017stable,ruthotto2020deep}. 
Since training the RNN model involves minimizing the prediction error. the choice of solver utilized to simulate the dynamic model during training would affect the resulting model quality. 
This is because the solver will introduce an additional numerical integration error when minimizing the cost function. 
Neural Ordinary Differential Equations (NODE), proposed by \cite{chen2018neuralode}, is an RNN model structure with continuous dynamics that can be simulated using conventional ODE solvers to evaluate the state variables at specific time steps but also to compute gradients when training the model. 
Designing the NODE model from physical insights can have a physically-based network structure without the need of limiting the implementation to a given integration method or step length. Still, \cite{zhu2022numerical} have shown that the NODE model will be an approximation of the true ODE, 
which is not only dependent on the quality of training data but also on the selected solver. More research is needed to understand the impact of using different numerical integration methods when training an RNN model. 

\subsection{Problem Statement}
The main objective of this work is to investigate how the choice of an ODE solver will affect the quality of a trained NODE model. 
Physically-based mathematical models often have few parameters to fit, where each parameter has a physical interpretation, which makes them robust against overfitting. 
On the other hand, neural networks can have millions of parameters; making them flexible but also susceptible to overfitting. Thus, the choice of ODE solver will have a direct impact on the properties of the trained model. 
In on-board fault diagnosis applications, where an accurate model is necessary but there are also requirements on computation time, it is important to understand the impact of an ODE solver on the implementation of a residual generator.

The system to be modeled can be described by a non-linear state-space model
\begin{equation} \label{eq:ODE_Model_example}
\begin{aligned}
\dot{x} &= g\left(x,u\right)\\
y &= h(x,u)\\
\end{aligned}
\end{equation}

where $x$ are state variables, $u$ are known input signals (e.g. actuator signals), $y$ are output signals (e.g. sensor outputs), and $g$ and $h$ are non-linear functions. 
Even though the system is continuous, the known signals $u$ and $y$ are sampled signals with a fixed sampling rate $T$. 
Thus, the trained model will be given sampled input data $u$ and is used to predict the measured output $y$ at the given time instances. 
While numerical solvers using adaptive step length are used in simulation tools to balance error tolerances and computation time, see e.g. \cite{ascher1998computer}, 
it is not desirable in online applications, such as residual evaluation, because of real-time requirements. Therefore, only explicit fixed-step solvers are considered in this study. 
A block diagram of the model is illustrated in Fig.~\ref{fig:Computational_graph}. 

\tikzset{
    vertex/.style={rectangle,draw,minimum width=6em},
    edge/.style={->,> = latex'},
    cvertex/.style={circle,draw,minimum width=0.3em,inner sep=1},
    lcvertex/.style={circle,draw,minimum width=2em,inner sep=1}}

\begin{figure}[!t]
\vspace{0.05in}
\centering
\begin{tikzpicture}[scale=1]
\input{comp_graph_mdl_tikz}
\end{tikzpicture}
\caption{An illustration of a block diagram of the nonlinear state-space model in (\ref{eq:ODE_Model_example}).}
\label{fig:Computational_graph}
\vspace{0.05in}
\end{figure}
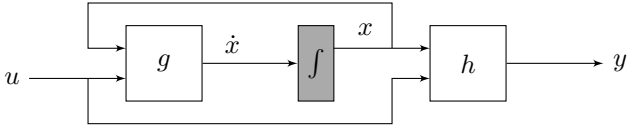


%% file: comp_graph_mdl_tikz.tex
\node (Input) at (-4,-0.2) { $u$};
\node[{rectangle,draw,minimum height = 1cm, minimum width = 1cm}] (Alg1) at (-2,0) {$g$};
\node (dh) at (-1.1,0.25) { $\dot{x}$};
\node[rectangle,draw,minimum height = 1cm,fill={rgb:black,1;white,2}] (ODE) at (0,0) { $\int$};
\node (h) at (0.65,0.45) { $x$};
\node[{rectangle,draw,minimum height = 1cm, minimum width = 1cm}] (Alg2) at (2,0) { $h$};
\node (Prediction) at (4,0) { $y$};
\draw[edge] (Input) -- (-2.5,-0.2);
\draw[edge] (1.0, 0.2) |- (-3, 0.8) -- (-3,0.2) -- (-2.5,0.2); 
\draw[edge] (0.23, 0.2) -- (1.5,0.2); 
\draw[edge] (-3,-0.2) |- (1, -0.8) -- (1,-0.2) -- (1.5,-0.2); 
\draw[edge] (Alg1) to[out=0,in=180] (ODE);
\draw[edge] (Alg2) to[out=0,in=180] (Prediction);

%% file: background.tex
\section{Modeling dynamic systems using Recurrent Neural Networks}
\label{sec:background}
The feed-forward neural network or multilayer perceptron (MLP) is perhaps the most well-known, and most widely employed network type when it comes to deep learning \citep{goodfellow2016deep}.
These models consist of (several) layers where each layer operation typically combines a linear combination of the inputs $x_{in,k}$ with a non-linear activation function $\phi$ as 
\begin{equation}
x_{out} =\phi(\Sigma_k W_k x_{in.k} + b),
\end{equation} 
where $W_k$ and $b$ are learnable parameters. 
In the case of dynamic systems where data is sequential, RNNs are generally more suitable. 
RNNs are time-discrete models and are not suitable to learn continuous functions such as \eqref{eq:ODE_Model_example}, but instead a time-discrete version:
\begin{equation} \label{eq:time-discrete}
\begin{aligned}
x_{k+T} &= \tilde{g}\left(x_k,u_k\right)\\
y_k &= \tilde{h}(x_k,u_k)\\
\end{aligned}
\end{equation}
where $k$ is the sample index, $T$ is sample time, and the functions $\tilde{g}$ and $\tilde{h}$ are non-linear functions modeled by the RNN. 
Note that $\tilde{g}$ also models the numerical integration of the state $x$ at the consecutive sample time $k+T$. 
RNN models are trained using back-propagation through time where gradients are computed using the chain rule \citep{werbos1990backpropagation}.

\subsection{RNN and Integration of Numerical Solver}

It is also possible to tailor the RNN structure so the model integrates the structure for a specific solver. 
For example in \cite{jung2022automated}, the structure of the RNN is designed to implement an EF solver to compute the states at time $k+T$ as
\begin{equation} \label{eq:GBNN_example}
\begin{aligned}
x_{k+T} &= x_k + T\hat{g}(x_k,u_k)\\
y_k &= \hat{h}(x_k,u_k)\\
\end{aligned}
\end{equation}  
where $\hat{g}$ and $\hat{h}$, ideally, will model the same system as in \eqref{eq:ODE_Model_example}. 
Similarly, other solvers can also be integrated into the network structure, such as the midpoint method (MP):
\begin{equation} \label{eq:midpoint_example}
\begin{aligned}
k_1 &= \hat{g}(x_k,u_k) \\
x_{k+T} &= x_k + T\hat{g}\left(x_k + Tk_1/2, u_{k + T/2}\right)\\
y_k &= \hat{h}(x_k,u_k)
\end{aligned}
\end{equation}  
where $T$ is the sample time and $u_{k+T/2} = \frac{1}{2}\left(u_k + u_{k+T}\right)$. 
Similarly, the corresponding model using the RK4 solver can be implemented as
\begin{equation} \label{eq:RK4_example}
\begin{aligned}
k_1 &= \hat{g}(x_k,u_k) \\
k_2 &= \hat{g}(x_k+Tk_1/2,u_{k+T/2}) \\
k_3 &= \hat{g}(x_k+Tk_2/2,u_{k+T/2}) \\
k_4 &= \hat{g}(x_k + Tk_3,u_{k+T}) \\
x_{k+T} &= x_k + \frac{T}{6}\left(k_1 + 2k_2 + 2k_3 + k_4\right)\\
y_k &= \hat{h}(x_k,u_k)
\end{aligned}
\end{equation}  
Note that the MP and RK4, in contrast to EF evaluate the model at points in between the sampling times, i.e., $k+T/2$. This can be done by, e.g., interpolating the model inputs.
  
\subsection{Neural Ordinary Differential Equations}

Instead of integrating the numerical integration method within the time-discrete RNN model, the authors in \cite{chen2018neuralode} propose the NODE model that directly models a time-continuous system such as \eqref{eq:ODE_Model_example}, see Fig.~\ref{fig:NODE_example_final}. 
The solutions can then be retrieved using an ODE solver. A wide range of solvers from fixed and adaptive step length to explicit and implicit options are available through the \texttt{torchdiffeq} library \citep{torchdiffeq} which is used in this work.

\pgfmathsetmacro{\nn}{4}

\begin{figure}
\centering
\begin{tikzpicture}[scale=1]

\tikzstyle{every pin edge}=[<-,shorten <=1pt]
\tikzstyle{neuron}=[circle,fill=black!30,minimum size=15pt,inner sep=5pt]
\tikzstyle{annot} = [text width=4em, text centered]

\node[neuron] (N111) at (0,0) {};
\node[neuron] (N112) at (0,1) {};
\node[neuron] (N121) at (0.8,-0.5) {};
\node[neuron] (N122) at (0.8,0.5) {};
\node[neuron] (N123) at (0.8,1.5) {};
\node[neuron] (N141) at (1.6,0.5) {};

\draw[edge] (N111) -- (N121);\draw[edge] (N111) -- (N122);\draw[edge] (N111) -- (N123);
\draw[edge] (N112) -- (N121);\draw[edge] (N112) -- (N122);\draw[edge] (N112) -- (N123);
\draw[edge] (N121) -- (N141);\draw[edge] (N122) -- (N141);\draw[edge] (N123) -- (N141);

\draw [dotted, thick, rounded corners] (-0.4,1.9) -- (2,1.9) -- (2,-0.9) -- (-0.4,-0.9) -- cycle;
\node (g) at (0.8,2.25) {$\hat{g}$};

\node[neuron] (N211) at (0+\nn,0) {};
\node[neuron] (N212) at (0+\nn,1) {};
\node[neuron] (N221) at (0.8+\nn,-0.5) {};
\node[neuron] (N222) at (0.8+\nn,0.5) {};
\node[neuron] (N223) at (0.8+\nn,1.5) {};
\node[neuron] (N241) at (1.6+\nn,0.5) {};

\draw[edge] (N211) -- (N221);\draw[edge] (N211) -- (N222);\draw[edge] (N211) -- (N223);
\draw[edge] (N212) -- (N221);\draw[edge] (N212) -- (N222);\draw[edge] (N212) -- (N223);
\draw[edge] (N221) -- (N241);\draw[edge] (N222) -- (N241);\draw[edge] (N223) -- (N241);

\draw [dotted, thick, rounded corners] (-0.4+\nn,1.9) -- (2+\nn,1.9) -- (2+\nn,-0.9) -- (-0.4+\nn,-0.9) -- cycle;
\node (h) at (0.8+\nn,2.25) {$\hat{h}$};

\node[rectangle,draw,minimum height = 1cm,fill={rgb:black,1;white,2}] (ODE) at (2.75,0.5) { $\int$};

\node (Input) at (-1.5,0.2) {$u$};
\node (y) at (1.6+\nn+ 1,0.5) {$y$};
\node (Imag) at (3.5,0.5) {};
\node (Imag_2) at (-0.25,0.5) {};
\node (Imag_3) at (1.75,0.75) {};

\draw[edge] (N141) -- (2.5,0.5);
\draw[edge] (N241) -- (y);

\draw[edge] (Input) -- (-0.4,0.2);
\draw[edge] (3.0,0.7) -- (3.6,0.7);

\draw[edge] (3.3, 0.7) |- (-1, 2.0) -- (-1,0.7) -- (-0.4,0.7); 

\draw[edge] (-1,0.2) |- (3.3, -1.0) -- (3.3,0.2) -- (3.6,0.2); 

\node (ODE_output) at (2.2,0.75) {$\dot{x}$};
\node (ODE_output) at (3.15,0.9) {$x$};

\end{tikzpicture}

\caption{An illustration of a NODE model of \eqref{eq:ODE_Model_example}.}
\label{fig:NODE_example_final}
\end{figure}
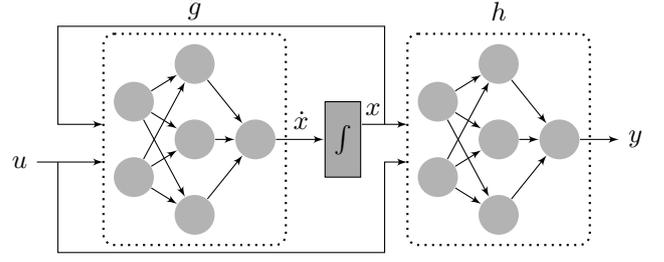

%% file: casestudy.tex
\section{Case study}
\label{sec:casestudy}
 
\begin{figure}[!b]
\vspace{0.1in}
	\begin{center}
		\includegraphics[width=1.0\columnwidth]{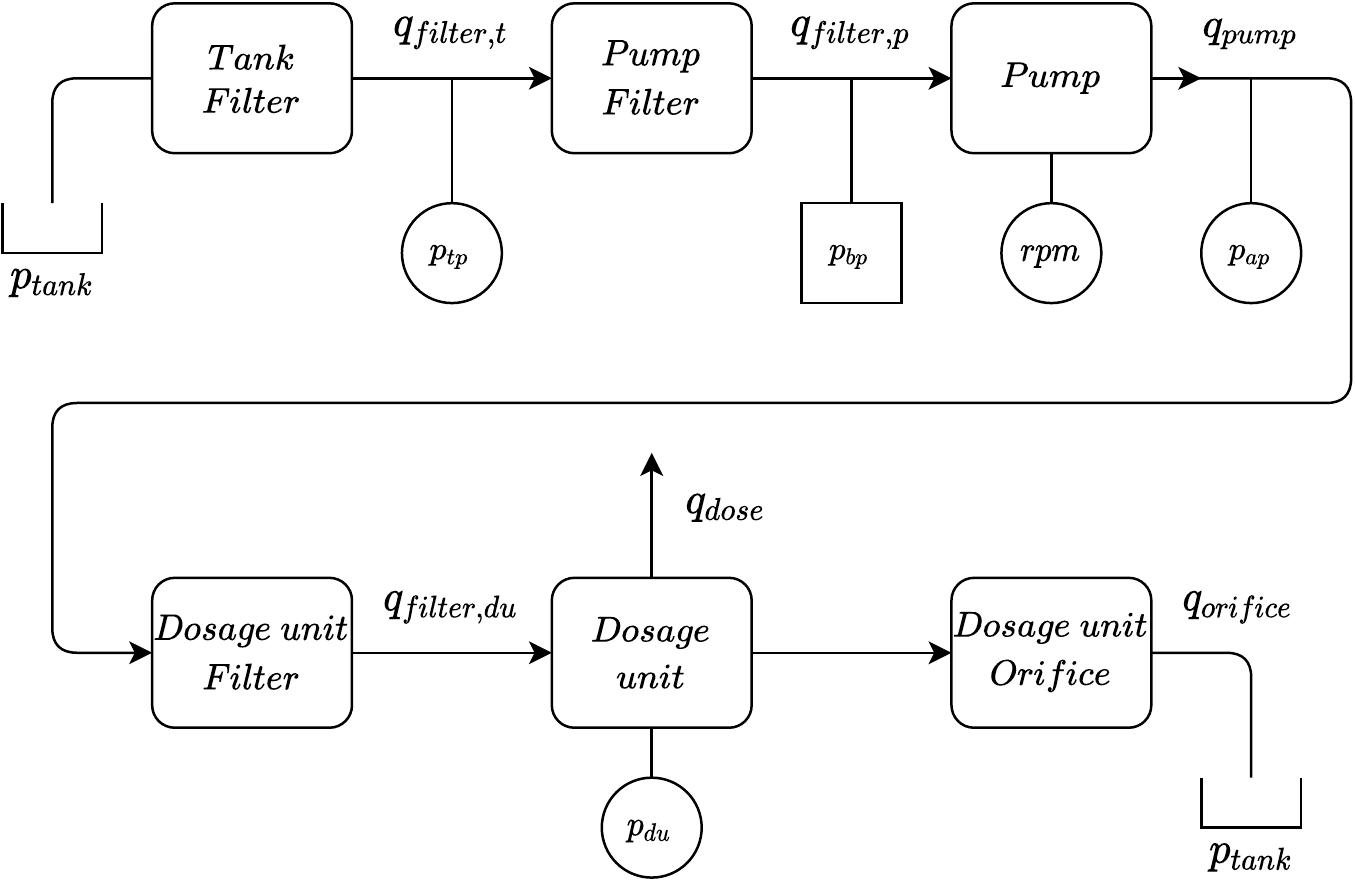}
		\caption{Schematic view of components in the after treatment system model \citep{jung2022fault}.} 
		\label{fig:schematic}
	\end{center}
\end{figure}

The system used as a case study is part of the exhaust gas after-treatment system in a heavy-duty truck \cite{jung2022automated}, as illustrated in Fig.~\ref{fig:schematic}. 
The system is responsible for dosing the appropriate amount of urea into exhausts to reduce NOx emissions.
The urea is stored in a tank from where it is pumped through a series of hoses to a dosage unit that injects the urea into the exhaust via a nozzle.
The dosing unit is controlled by a PWM signal.
The available signals are summarized in Table~\ref{tb:signals}.
Note that the duty cycle only gives the intensity of the PWM signal controlling the dosing unit and not the phase of the actual control signal \citep{jung2022fault}.
A model of the after-treatment system that is presented here is based on the work in \cite{jung2022fault} with some adjustments. The system dynamics are modeled by the state variables governing the pressures before and after the pump and the pressure inside the dosing unit. 

\begin{table}[!t]
	\caption{Available signals.}
	\label{tb:signals}
	\centering
	\begin{tabular}{c l}
		\toprule
		Signal & Description\\
		\midrule
		$y_{p,tp}$ & Pressure before the pump filter\\ 
		$y_{p,ap}$ & Pressure before after pump filter\\
		$y_{p,du}$ & Pressure inside the dosing unit\\		
		$n_p$ & Pump speed\\
		$DC$ & Duty cycle of the dosing unit\\
		\bottomrule
	\end{tabular}
\end{table}

Three different RNN-based residuals are implemented using NODE in this case study to be able to switch between different solvers.
The RNN-based residuals are derived from a structural model that has been formulated based on the model relations presented in \cite{jung2022fault}.
Even though it is not the focus of this work, the motivation of this design approach is to make the residual generator based on the RNN model, sensitive to certain faults, even though only fault-free training data is available, which is useful for the isolation of unknown faults \citep{jung2019isolation}. 

\begin{figure}[!t]
	\begin{center}
		\includegraphics[width=1.0\columnwidth]{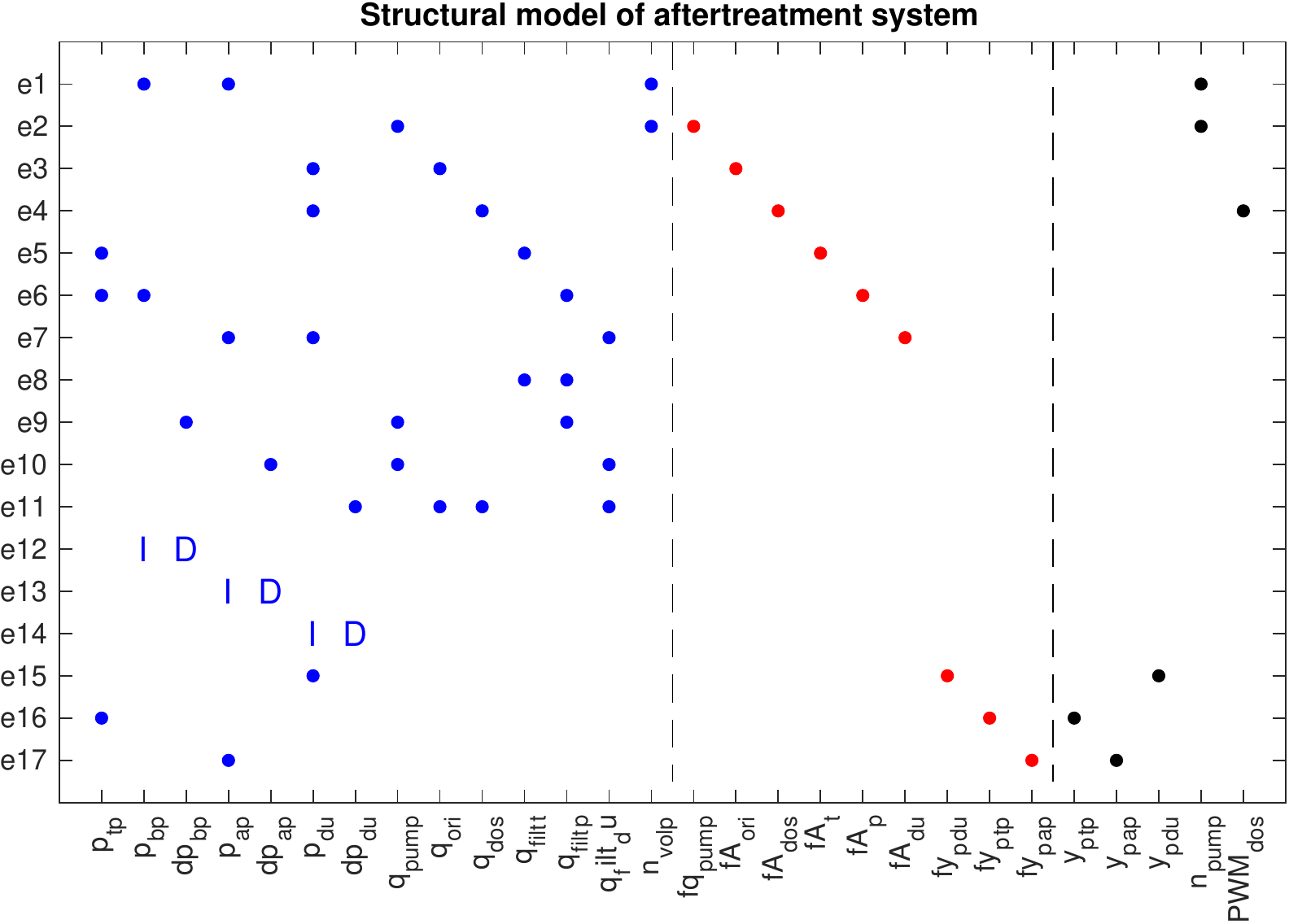}    
		\caption{Structural model of the after-treatment system. The chart uses blue dots for unknown variables, red dots for faults, 
		and black dots for known variables on the horizontal axis. } 
		\label{fig:Str_mod}
	\end{center}
\end{figure}

The structural model is a bipartite graph that describes the relations between equations and model variables. Fig.~\ref{fig:Str_mod} shows the equations, the unknown variables including the three dynamic states (marked I) and their derivatives (marked D), faults, and known variables. 
The three selected RNN models are each containing one, two, and three dynamic states, respectively. The RNN models are used to construct three different residual generators, denoted $r_1$, $r_2$, and $r_3$. 

\subsection{Design of RNN-based Residual Generators}
\label{sec:design}

From the structural model, three computational graphs are derived to construct each of the RNN model structures. 
Fig.~\ref{fig:Computational graph} shows one computational graph and the influence of each fault 
in the creation of the above-mentioned residual. The blue-colored variables are the measured signals and the red-colored variables are faults.

\tikzset{
    vertex/.style={rectangle,draw,minimum width=6em},
    edge/.style={->,> = latex'},
    cvertex/.style={circle,draw,minimum width=0.3em,inner sep=1},
    lcvertex/.style={circle,draw,minimum width=2em,inner sep=1}    
}

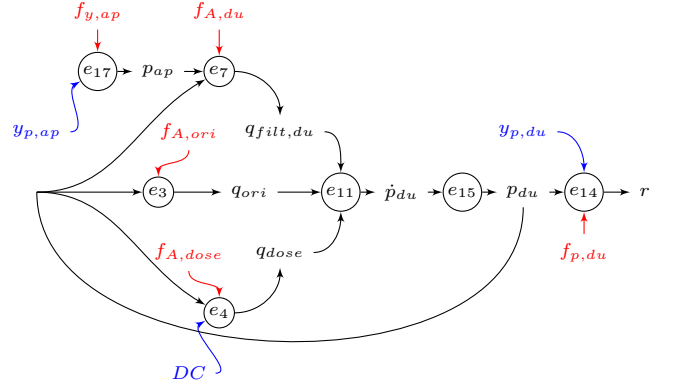
\begin{figure}[!h]
\centering
\begin{tikzpicture}[scale=0.8]
\input{comp_graph_design_tikz}
\end{tikzpicture}
    \caption{An illustration of a computational graph used to design the RNN models.}
    \label{fig:Computational graph}
\end{figure}

The grey-box RNN-based residual generation models the known relations between input and output signals, based on the computational graph, similarily to model-based residuals but the analytical relationships are considered unknown. This approach constructs NODE networks that respect the known dependencies between state variables and measurements. The resulting NODE model from the computational graph in Fig.~\ref{fig:Computational graph} can be formulated in state-space form as 
\begin{equation} \label{eq:DAE}
\begin{aligned}
\dot{p}_{du} &= \hat{g}_{1}({p}_{du}, y_{p,ap}, DC) \\
r_t &= \hat{h}_{1}(p_{du}) - y_{p,du}
\end{aligned}
\end{equation}
Two more residuals following the same pattern are generated, the second has two dynamic states as follows:
\begin{equation} \label{eq:MSO6}
\begin{aligned}
\begin{pmatrix}\dot{p}_{ap} \\ \dot{p}_{bp} \end{pmatrix} &= \begin{pmatrix} \hat{g}_{21}({p}_{ap}, {p}_{bp}, y_{p,du}, n_p) \\ \hat{g}_{22}({p}_{ap}, {p}_{bp}, y_{p,tp}, n_p) \end{pmatrix} \\
r_{2} &= \hat{h}_2(p_{ap}) - y_{p,ap}
\end{aligned}
\end{equation}
Finally, the third residual uses all three dynamic states in the structure as given by:
\begin{equation} \label{eq:MSO13}
\begin{aligned}
 \begin{pmatrix} \dot{p}_{du} \\ \dot{p}_{ap} \\ \dot{p}_{bp} \end{pmatrix} &= \begin{pmatrix} \hat{g}_{31}({p}_{du}, y_{p,ap}, DC) \\ \hat{g}_{32}({p}_{ap}, {p}_{bp}, {p}_{du}, n_p) \\ \hat{g}_{33}({p}_{ap}, {p}_{bp}, y_{p,tp}, n_p) \end{pmatrix} \\
r_{3} &= \hat{h}_3(p_{du}) - y_{p,du}
\end{aligned}
\end{equation}


\subsection{Data collection and training}
\begin{figure}[!h]
\begin{center}
\includegraphics[width=1.0\columnwidth]{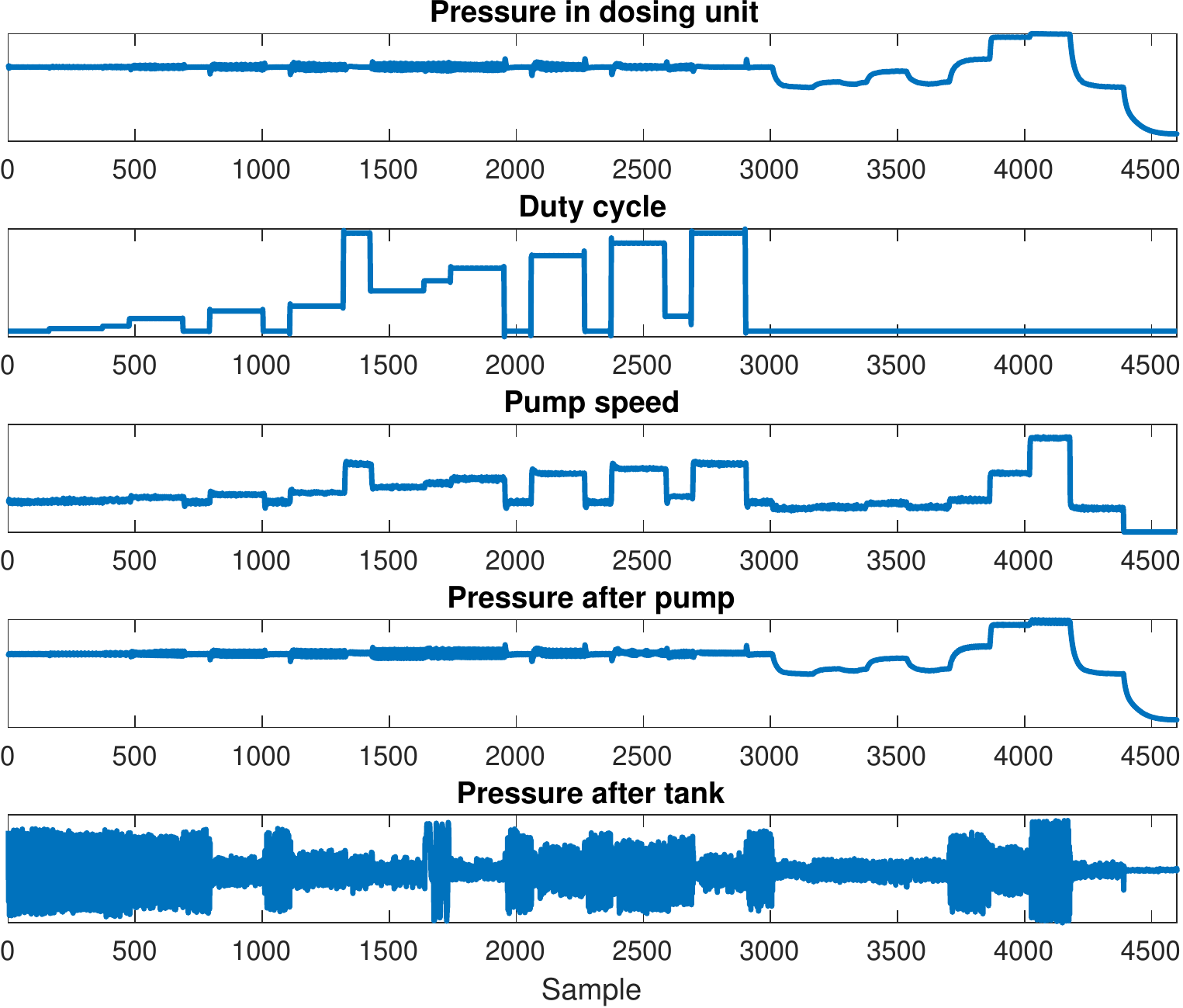}    
\caption{Example of collected data under nominal operating conditions.} 
\label{fig:NF_data}
\end{center}
\end{figure}

Collected data from a heavy-duty truck is provided to evaluate the performance of the diagnosis system for different fault scenarios. Besides two fault-free data sets, one for training and one for testing, data have been collected from fault scenarios including clogging before the dosing unit $f_{A,du}$, clogging after the dosing unit $f_{A,ori}$, and clogging after the pump $f_{A,p}$. Figure~\ref{fig:NF_data} shows the fault-free training data set of 4600 samples used in this work. The validation data consist of 2300 samples of fault-free data, as well as 2300 samples of each faulty case scenario.

The same neural network structure is used to learn each function $\hat{g}_i$ and $\hat{h}_j$.
The network consists of two hidden layers with a dimension of 128 and ELU activations.
(while ELU is slower in convergence than ReLU but can handle negative input without dying out) 
The model parameters are learned by minimizing the Huber-loss of the prediction error (chosen due to fast initial convergence) with Adam \citep{kingma2014adam} as optimizer. 
To evaluate the intermediate steps of the Midpoint (MP) method \eqref{eq:midpoint_example} and the Runge-Kutta method of order four (RK4) \eqref{eq:RK4_example}, the model input signals are computed using linear interpolation.
During training, the samples that make up the mini-batches are a random selection sequence of 400 consecutive samples. 
The initial value for the state variables of each sample is drawn from a normal distribution with mean and variance based on estimates of the measurement signals in the training set.
Since these estimates may not be available for all residuals, this approach is an attempt to make the model robust against poorly chosen initial states.
During inference, the estimated mean is taken directly as the initial state for the sequence.

%% file: comp_graph_design_tikz.tex

\node[blue] (y_pap) at (0,0) {\scriptsize $y_{p,ap}$};

\node[cvertex] (e1) at (1,1) {\scriptsize $e_{17}$};

\node[red] (f_yap) at (1,2) {\scriptsize $f_{y,ap}$};

\node (p_ap) at (2,1) {\scriptsize $p_{ap}$};

\node[cvertex] (e2) at (3,1) {\scriptsize $e_7$};

\node[red] (f_Adu) at (3,2) {\scriptsize $f_{A,du}$};

\node[cvertex] (e4) at (2,-1) {\scriptsize $e_3$};

\node[red] (f_Aori) at (2.5,0) {\scriptsize $f_{A,ori}$};

\node[cvertex] (e3) at (3,-3) {\scriptsize $e_4$};

\node[red] (f_Adose) at (2.5,-2) {\scriptsize $f_{A,dose}$};

\node[blue] (DC) at (2.5,-4) {\scriptsize $DC$};

\node (q_filtdu) at (4,0) {\scriptsize $q_{filt,du}$};

\node (q_ori) at (3.5,-1) {\scriptsize $q_{ori}$};

\node (q_dose) at (4,-2) {\scriptsize $q_{dose}$};

\node[cvertex] (e5) at (5,-1) {\scriptsize $e_{11}$};

\node (dp_du) at (6,-1) {\scriptsize $\dot{p}_{du}$};

\node[cvertex] (e6) at (7,-1) {\scriptsize $e_{15}$};

\node (p_du) at (8,-1) {\scriptsize $p_{du}$};

\node[blue] (y_pdu) at (8,0) {\scriptsize $y_{p,du}$};

\node[cvertex] (e7) at (9,-1) {\scriptsize $e_{14}$};

\node[red] (f_pdu) at (9,-2) {\scriptsize $f_{p,du}$};

\node (r) at (10,-1) {\scriptsize $r$};


\draw[blue, edge] (y_pap) to[out=0,in=210] (e1);

\draw[red, edge] (f_yap) to[out=270,in=90] (e1);

\draw[edge] (e1) to[out=0,in=180] (p_ap);

\draw[edge] (p_ap) to[out=0,in=180] (e2);

\draw[red, edge] (f_Adu) to[out=270,in=90] (e2);

\draw[edge] (e2) to[out=0,in=90] (q_filtdu);

\draw[edge] (e4) to[out=0,in=180] (q_ori);

\draw[red, edge] (f_Aori) to[out=270,in=90] (e4);

\draw[edge] (e3) to[out=0,in=270] (q_dose);

\draw[red, edge] (f_Adose) to[out=270,in=90] (e3);

\draw[blue, edge] (DC) to[out=0,in=210] (e3);

\draw[edge] (q_filtdu) to[out=0,in=90] (e5);

\draw[edge] (q_ori) to[out=0,in=180] (e5);

\draw[edge] (q_dose) to[out=0,in=270] (e5);

\draw[edge] (e5) to[out=0,in=180] (dp_du);

\draw[edge] (dp_du) to[out=0,in=180] (e6);

\draw[edge] (e6) to[out=0,in=180] (p_du);

\draw (p_du) to[out=270,in=270] (0,-1);

\draw[edge] (p_du) to[out=0,in=180] (e7);

\draw[red, edge] (f_pdu) to[out=90,in=270] (e7);

\draw[edge] (0,-1) to[out=0,in=210] (e2);

\draw[edge] (0,-1) to[out=0,in=180] (e4);

\draw[edge] (0,-1) to[out=0,in=150] (e3);

\draw[blue, edge] (y_pdu) to[out=0,in=90] (e7);

\draw[edge] (e7) to[out=0,in=180] (r);

\newcommand\rtwoy{-2.5}

\newcommand\rtwox{2.5}

%% file: evaluation.tex
\section{Evaluation}
\label{sec:Analysis}

In this section, the three RNN-based residuals described in Section \ref{sec:design} will be used as a case study. In the analysis, three different solvers are used to evaluate the RNN-models: EF, MP, and RK4.

\subsection{Training and validation performance of RNN models}
\begin{figure}
\begin{center}
\includegraphics[width=1.0\columnwidth]{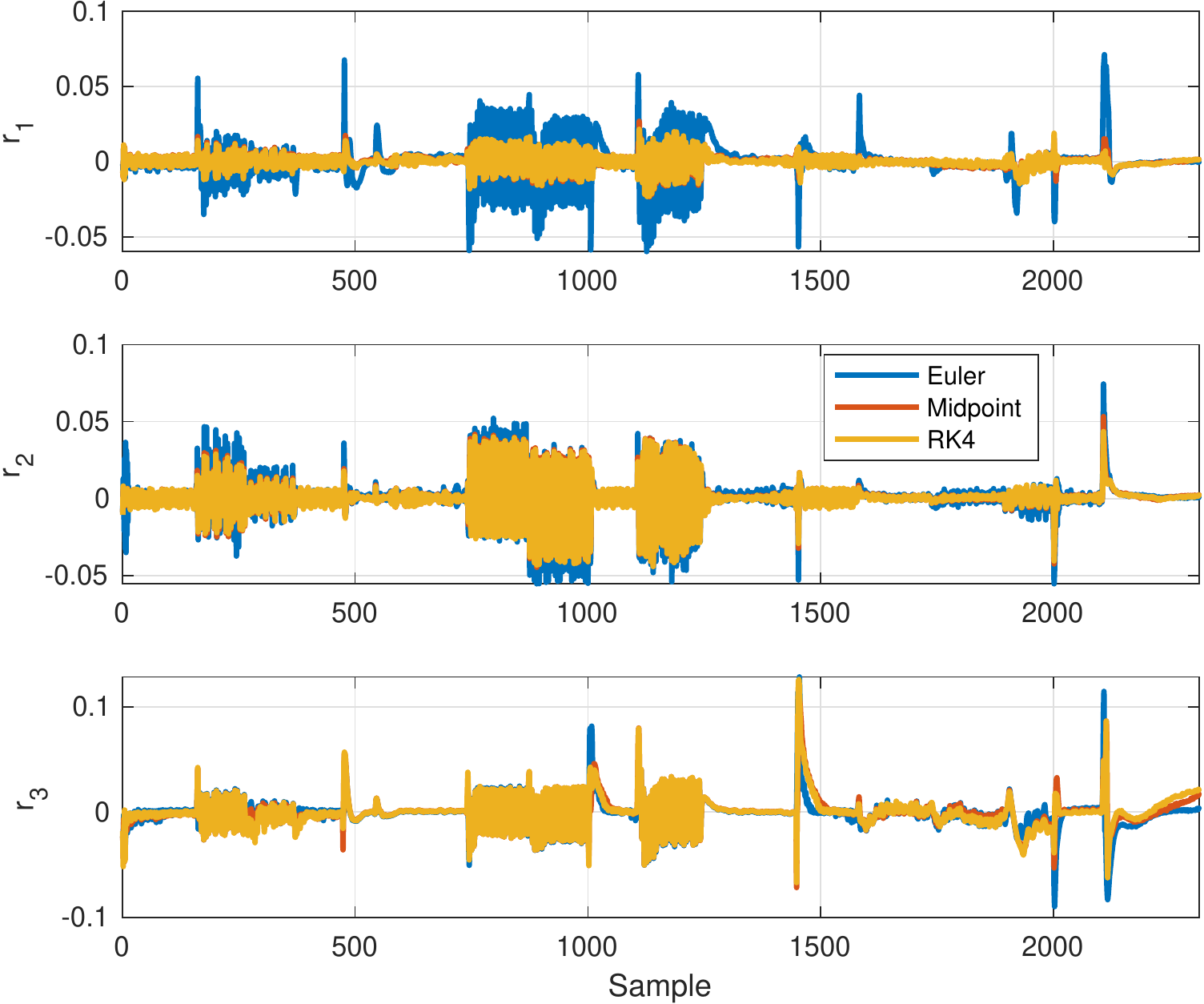}
\caption{Prediction performance of the three RNN-based models on nominal test data using different solvers.} 
\label{fig:Evaluation_on_NF_test}
\end{center}
\end{figure}

The residuals have been trained using the data shown in Fig.~\ref{fig:NF_data}, 
where the input signals are following the structure of each RNN model and the validation is performed using an independent nominal test set.
The prediction performances of the three RNN models, when trained using different solvers, are depicted in Fig.~\ref{fig:Evaluation_on_NF_test}.
While the general behavior of the signals is captured by the models, some oscillations occur due to the PWM signal controlling the dosing unit \citep{jung2022fault}. 
As a result, the validation loss is calculated based on the second half of the dataset when the dosing unit is closed. 
Each RNN model is trained multiple times and the best result for each model and solver combination is shown in Table~\ref{tb:training_summary}. 



\begin{table}[hb]
	\begin{center}
		\caption{Results from training the RNN models using different solvers.}
		\label{tb:training_summary}
		\begin{tabular}{ccccccc}
			\toprule
				 & \multicolumn{2}{c}{EF} &  \multicolumn{2}{c}{MP} & \multicolumn{2}{c}{RK4} \\
				RNN & Train & Val &  Train & Val & Train & Val \\
				\midrule
				$r_1$ & 5.1e-5 & 7.8e-5 & 1.3e-5 & 0.9e-5  & 1.2e-5 & 0.8e-5\\
				$r_2$ & 1.0e-5 & 4.5e-5 & 1.5e-5 & 2.1e-5  & 1.6e-5 & 1.7e-5\\
				$r_3$ & 5.8e-5 & 19.4e-5 & 5.6e-5 & 12.6e-5 & 6.2e-5 & 13.8e-5 \\
			\bottomrule
		\end{tabular}
	\end{center}
\end{table}

\begin{figure}
\begin{center}
\includegraphics[width=1.0\columnwidth]{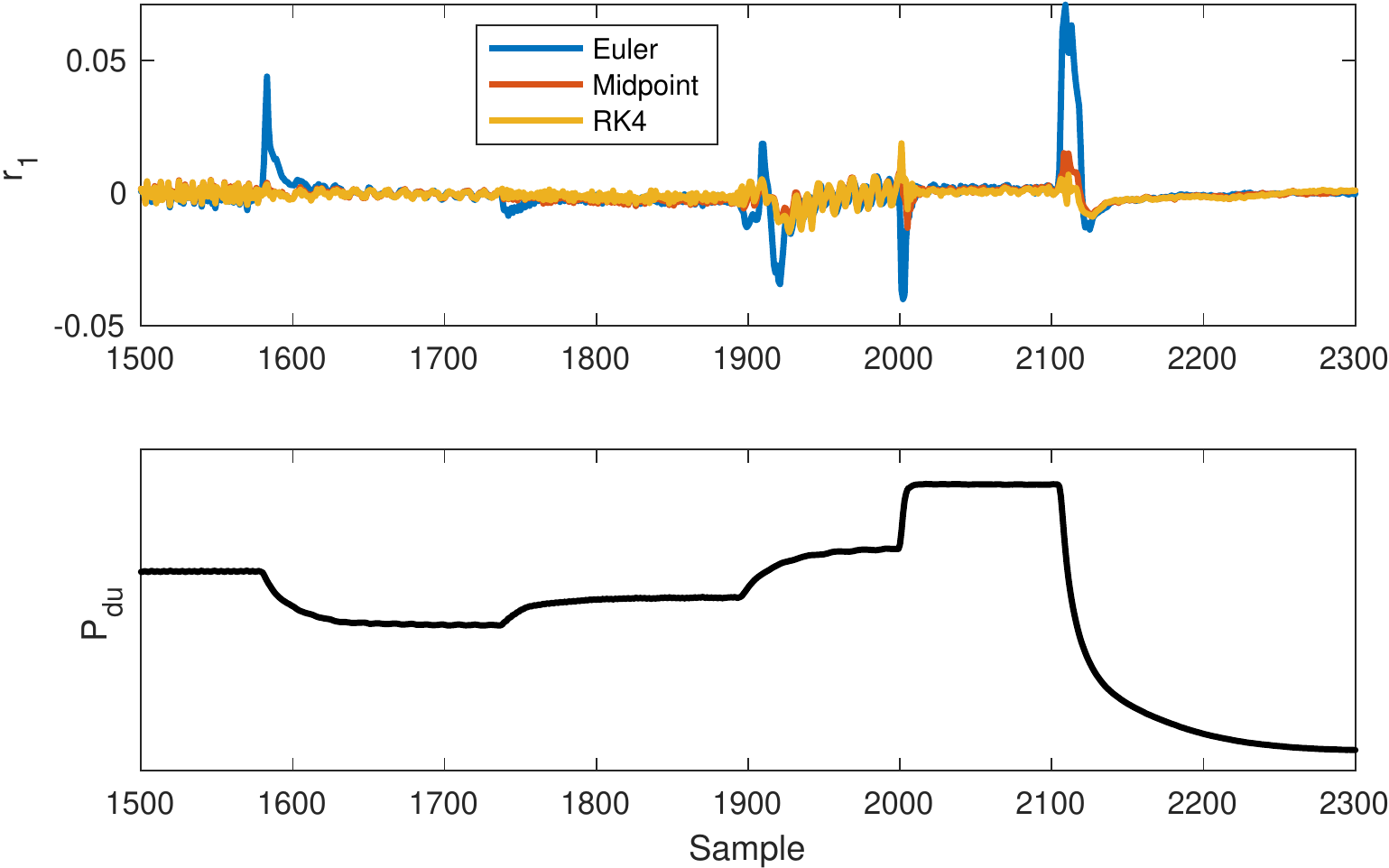}
\caption{Solvers performance in prediction of data no fault test set in the case of transient response} 
\label{fig:Bumps}
\end{center}
\end{figure}

To clarify the difference between the RNN models, Fig.~\ref{fig:Bumps} shows a zoomed-in interval of residual $r_1$ together with its referenced signal. 
The main difference between the methods is prediction performance during the transients.
The EF solver cannot follow the transients as accurately as MP and RK4, illustrating the benefit of using a higher-order method during training.

\subsection{Model Evaluation Using Different Solvers}
\begin{figure}
\begin{center}
\includegraphics[width=1.0\columnwidth]{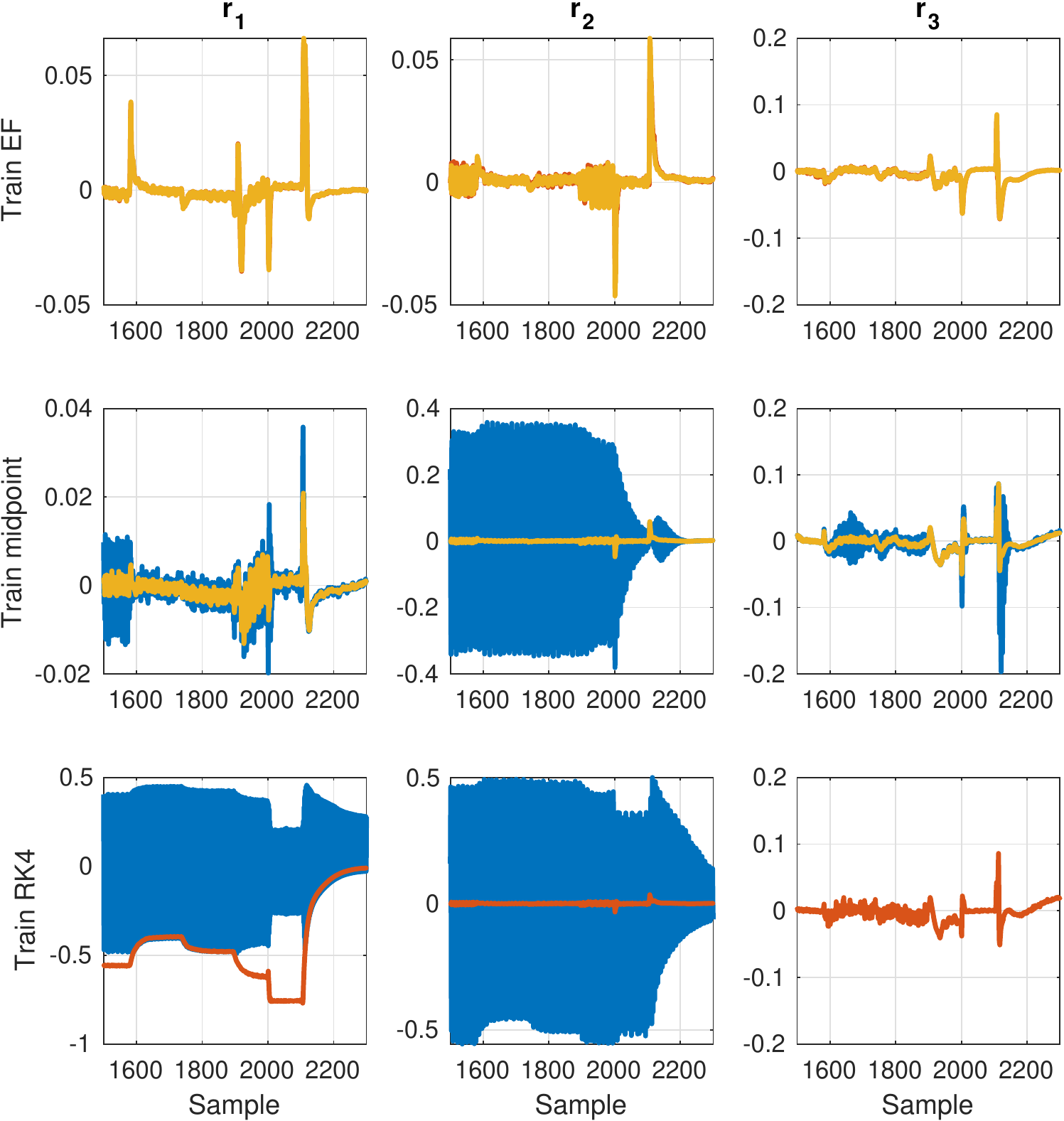}
\caption{Evaluation of three trained residuals evaluated on other solvers. 
The label on the vertical axis shows the solver method used during training and the color shows the solver used during evaluation. 
Blue curves are evaluated using EF, red curves are MP, and yellow curves are RK4. In the lower right curve, the evaluation using EF has diverged.
}  
\label{fig:Evaluation_Solver_change}
\end{center}
\end{figure}

All models capture the behavior of the system. The next step is to evaluate how the trained model generalize toward solver selection. This is done by selecting a model trained using one solver and evaluated it using another to determine if the network has learned the same underlying dynamics. 
Figure~\ref{fig:Evaluation_Solver_change} illustrates the output of the different RNN models when trained and evaluated using different solvers. In this plot, the test data is evaluated when the dosing unit is turned off. Each subplot depicts the residual outputs of the same trained model evaluated using the other two solvers. The corresponding mean squared error values are shown in Table~\ref{tb:change_solvers}. 


\begin{figure}
\begin{center}
\includegraphics[width=1.0\columnwidth]{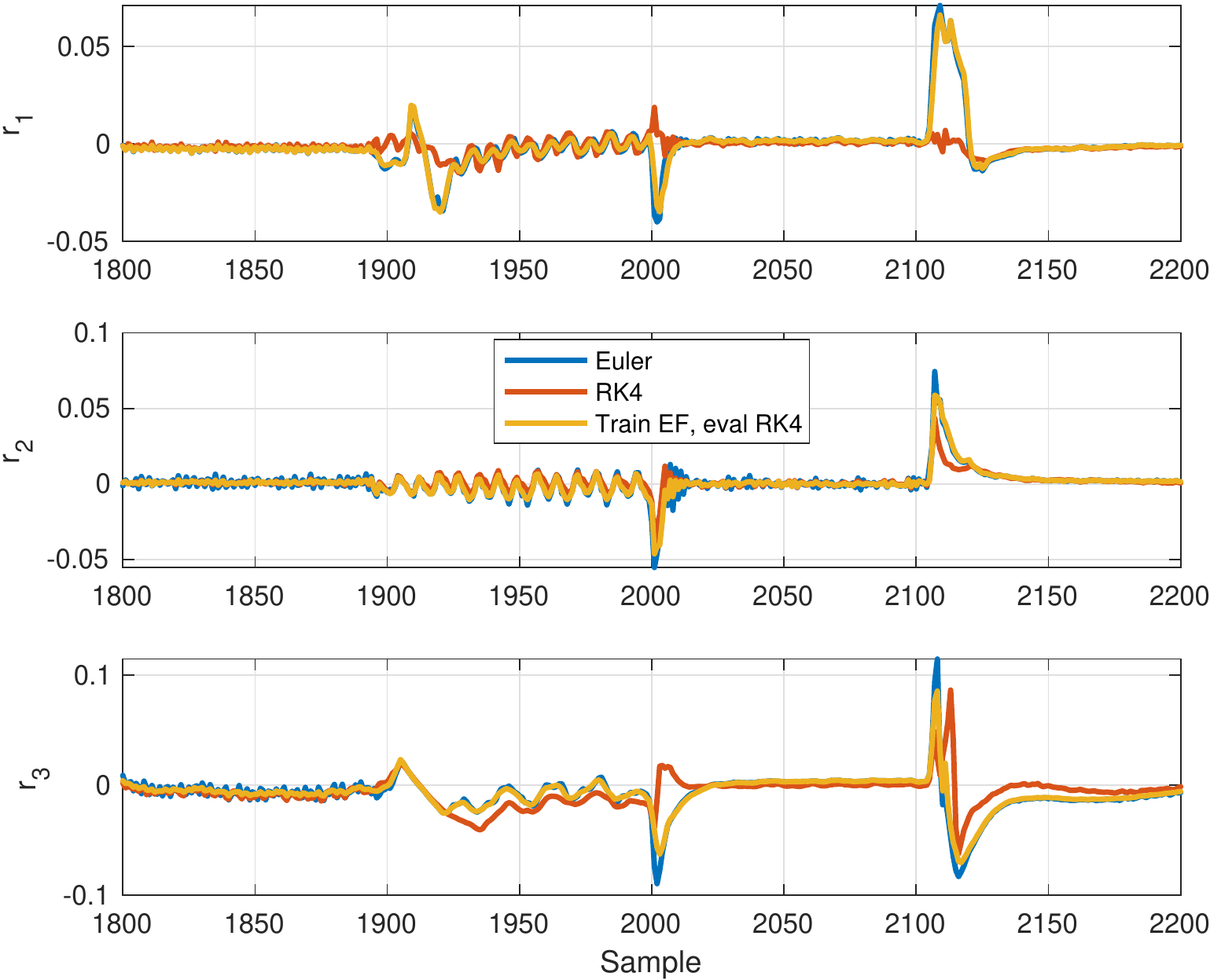}
\caption{Comparison of different integration methods with a case of a model trained on lower order and evaluated on higher one for three residuals.} 
\label{fig:Sol_up}
\end{center}
\end{figure}

{\setlength{\tabcolsep}{5pt}
\begin{table}[hb]
\begin{center}
\caption{Prediction error results from evaluating each of the RNN models using different solvers.}\label{tb:change_solvers}
\begin{tabular}{cccc}
	\toprule
      & \multicolumn{3}{c}{Trained EF}\\
Eval      & EF      & MP      & RK4 \\
\midrule
$r_1$ & 7.8e-5 & 7.0e-5 & 7.2e-5 \\
$r_2$ & 4.5e-5 & 3.3e-5 & 3.9e-5 \\
$r_3$ & 1.9e-4 & 1.5e-4 & 1.5e-4 \\
\bottomrule
\end{tabular}\hfill%
\begin{tabular}{cccc}
		\toprule
      & \multicolumn{3}{c}{Trained MP}\\
Eval      & EF      & MP      & RK4 \\
\midrule
$r_1$ & 2.1e-5 & 8.7e-6 & 8.6e-6 \\
$r_2$ & 0.4e-1 & 2.1e-5 & 2.4e-5 \\
$r_3$ & 4.1e-4 & 1.3e-4 & 1.3e-4 \\
\bottomrule
\end{tabular}

\vspace{0.2cm}

\begin{tabular}{cccc}
		\toprule
      & \multicolumn{3}{c}{Trained RK4}\\
Eval      & EF      & MP      & RK4 \\
\midrule
$r_1$ & 1.4e-1 & 2.4e-1 & 8.0e-6 \\
$r_2$ & 1.1e-1 & 1.5e-5 & 1.7e-5 \\
$r_3$ & - & 1.5e-4 & 1.4e-4 \\
\bottomrule
\end{tabular}
\end{center}
\end{table}
}
The first observation is when using RNN models that are trained using a lower-order solver and evaluating them using a higher-order one, that the prediction accuracy is not significantly improved. 
One possible explanation for this phenomenon is that the trained model becomes overfitted to a specific solver. 
This is consistent with the findings presented in \cite{zhu2022numerical}. 
However, other factors may also contribute to the lack of improvement. 
For instance, the stability regions of different solvers may limit the learning process of the model parameter which prevents the enhancement in prediction accuracy when using higher-order solvers.
This is illustrated in Figure~\ref{fig:Sol_up} which shows the outputs from the RNN models trained and evaluated using the same solver, EF and RK4, respectively. It also shows the case where the model is trained using EF and then evaluated using RK4. When $r_1$ is trained using EF and evaluated using RK4 the the prediction error at sample 2100 when using EF for both training and simulation is not improved when simulating using RK4. This indicates that the properties of the model are limited by the use of EF during training.

\begin{figure}
\begin{center}
\includegraphics[width=1.0\columnwidth]{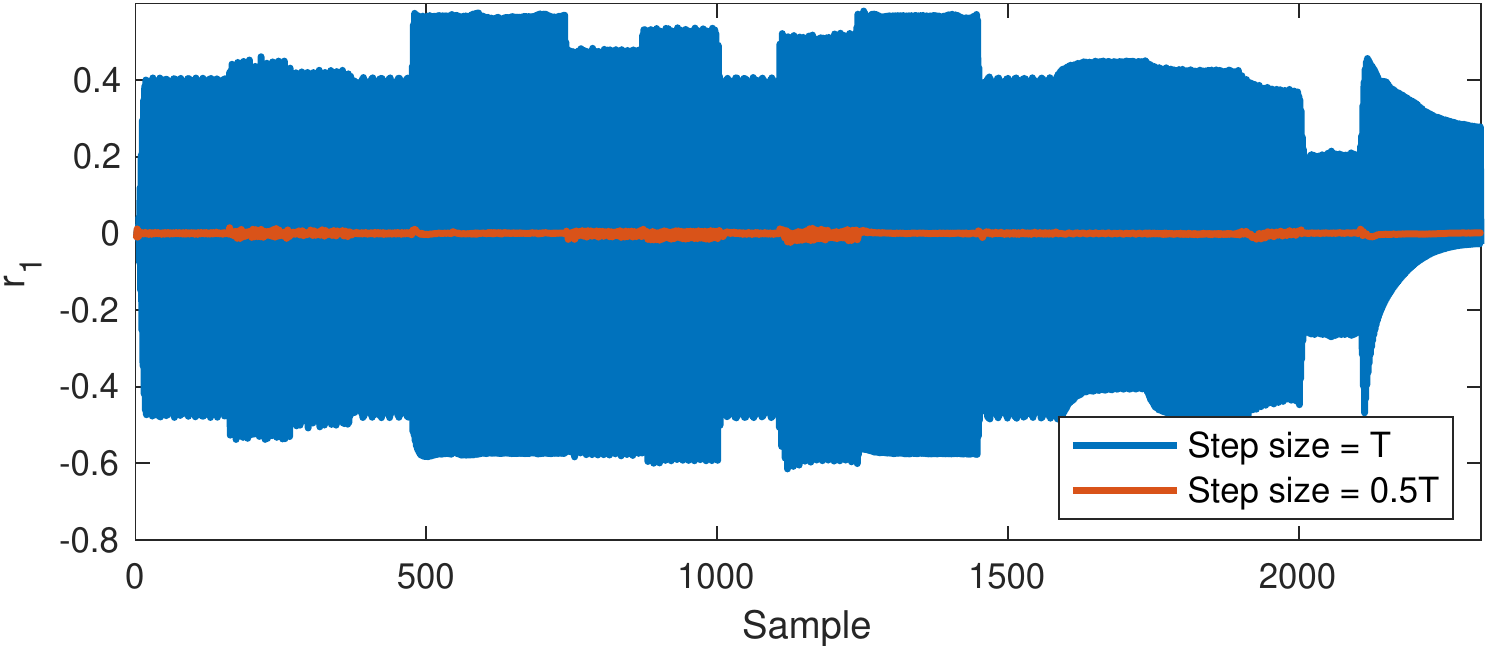}
\caption{Comparison of models trained with RK4 and evaluated on EF using different step sizes. The blue curve is using the same step size $T$ as during training while the red is using step size $0.5T$.} 
\label{fig:Sample_time}
\end{center}
\end{figure}

However, in evaluating RNN models trained with higher-order methods using lower-order solvers, a significant decrease in performance and occasional instability has been observed. 
One possible explanation for this is the larger stability region of higher-order solvers. 
When a model is trained using a higher-order solver, 
the poles of the trained model may be located within the stability region of the higher-order solver but outside the stability region of the lower-order one. 
This could result in an unstable model and poor performance \citep{ascher1998computer}. 
A solution to imporve stability is to reduce step size to bring the evaluation into the stability region.
Figure~\ref{fig:Sample_time} shows the output of residual $r_1$ when it is trained using RK4 and evaluated using EF with different time steps. 
Reducing the step size stabilizes the model predictions even though the model is trained using a higher-order solver.

\subsection{Fault Diagnosis}

\begin{figure}
\begin{center}
\includegraphics[width=1.0\columnwidth]{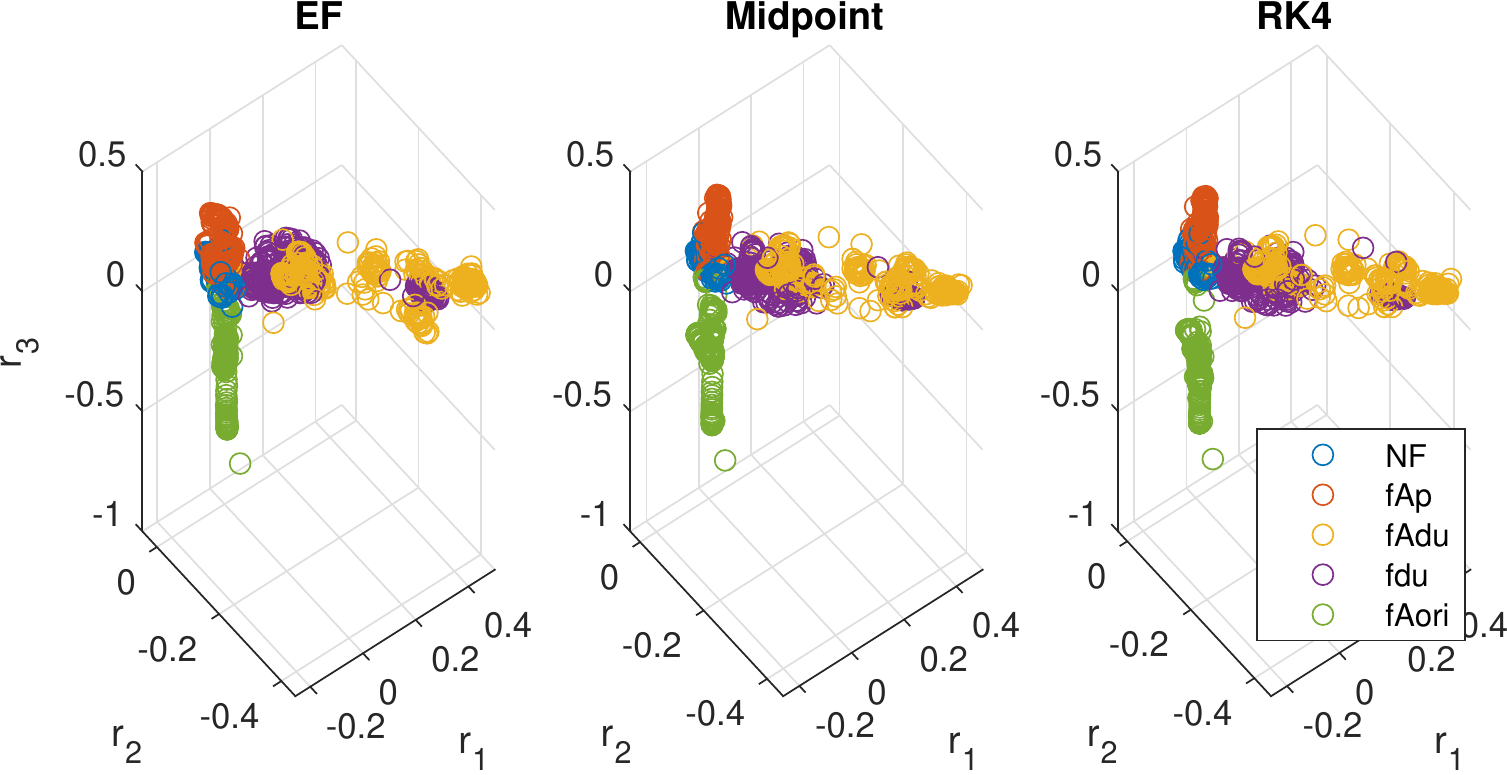}
\caption{Outputs from the three RNN-based residuals using the EF, midpoint, and RK4, methods, respectively, evaluated on different fault scenarios.} 
\label{fig:Class_all}
\end{center}
\end{figure}

In the final analysis, the RNN models are evaluated using data from different fault scenarios.
Here, the RNN models are evaluated using the same solvers as they were trained on.
The residual outputs from the three RNN models are plotted against each other in Fig.~\ref{fig:Class_all}. Each color represents samples from a specific fault scenario. It is visible that all faults are distinguishable from the nominal class in all cases and that the different fault classes are similarly distributed in residual space for all solvers. Since higher-order solvers, in general, performs better to simulate fast transients, and the dynamics in the data sets are limited, a significant change in detection performance were not expected. However, further analysis should be done to investigate the impact of varying noise levels due to the PWM-based control of the dosing unit. Still, reduncing the spikes in the residual outputs in Fig.~\ref{fig:Sol_up} during transients would likely reduce the risk of false alarms.

%% file: conclusions.tex
\section{Conclusions and Future Work}
The selection of a numerical integration method to simulate RNN models has a significant impact on the performance of the trained model. 
While a higher-order solver seems to give a better prediction accuracy both during training and simulation, the model performance will degrade significantly when changing to a lower-order evaluation method.
Thus, training with a lower-order method appears to result in a more robust model, at the cost of accuracy, especially during transients. 
Still, when training the RNN models, the selection of a solver doesn't seem to have an impact on the data distribution in the residual space during the evaluated fault scenarios. 
This is desirable from a fault diagnosis perspective since the selection should mainly affect model accuracy, i.e., the detection performance, and not the distribution of residual data from the different fault scenarios.

In future works, the analysis will include other types of numerical solvers, e.g., implicit and adaptive step length methods, to investigate their effects on the quality of the learned parameters of the underlying dynamic.
It is also relevant to conduct a deeper analysis on the relation of the stability region of numerical integration methods and the stability of RNN models.